\documentclass{article}

\usepackage{arxiv}
\usepackage{color} 
\usepackage[utf8]{inputenc} 
\usepackage[T1]{fontenc}    
\usepackage{hyperref}       
\usepackage{url}            
\usepackage{booktabs}       
\usepackage{amsfonts}       
\usepackage{nicefrac}       
\usepackage{microtype}      
\usepackage{lipsum}
\usepackage{amssymb}
\usepackage{amsmath}
\usepackage{graphicx}
\usepackage{multirow}
\usepackage{makecell}
\usepackage{algorithm}
\usepackage{algorithmic}
\newtheorem{theorem}{Theorem}

\newtheorem{Proof}{Proof}
\graphicspath{ {./images/} }

\title{Filter Grafting for Deep Neural Networks: Reason, Method, and Cultivation}

\author{Hao~Cheng$^{1*}$, Fanxu~Meng$^{1*}$,  Ke~Li$^1$, Yuting~Gao$^1$, Guangming Lu$^{2}$, Xing Sun$^{1\dagger}$,Rongrong~Ji$^{3}$\\ 
	$^1$ Tencent Youtu Lab, Shanghai, China \\
	$^2$ Harbin Institute of Technology, Shenzhen, China\\		
	$^3$ XiaMen University, XiaMen, China.\\	
}

\begin{document}
\maketitle
\begin{abstract}
	Filter is the key component in modern convolutional neural networks (CNNs). However, since CNNs are usually over-parameterized, a pre-trained network always contain some invalid (unimportant) filters. These filters have relatively small $l_{1}$ norm and contribute little to the output (\textbf{Reason}). While filter pruning removes these invalid filters for efficiency consideration, we tend to reactivate them to improve the representation capability of CNNs. In this paper, we introduce filter grafting (\textbf{Method}) to achieve this goal. The activation is processed by grafting external information (weights) into invalid filters. To better perform the grafting, we develop a novel criterion to measure the information of filters and an adaptive weighting strategy to balance the grafted information among networks. After the grafting operation, the network has fewer invalid filters compared with its initial state, enpowering the model with more representation capacity. Meanwhile, since grafting is operated reciprocally on all networks involved, we find that grafting may lose the information of valid filters when improving invalid filters. 
	To gain a universal improvement on both valid and invalid filters, we compensate grafting with distillation (\textbf{Cultivation}) to overcome the drawback of grafting . Extensive experiments are performed on the classification and recognition tasks to show the superiority of our method. Code is available at \textcolor{black}{\emph{https://github.com/fxmeng/filter-grafting}}.
\end{abstract}

\footnotetext[1] {In the author list, $^{\ast}$ denotes that authors contribute equally; $^{\dagger}$ denotes corresponding authors. }

\section{Introduction}
For the past decade, Deep Neural Networks (DNNs) have demonstrated their great successes in many research areas, including computer vision \cite{krizhevsky2012imagenet,lotter2016deep}, speech recognition \cite{graves2013speech}, and language processing \cite{zhang2015text}. However, recent studies show that DNNs have invalid (unimportant) filters \cite{li2016pruning}. These filters have little effect on the output accuracy and removing them could accelerate the inference of DNNs without hurting much performance. This discovery inspires many works to study how to identify unimportant filters precisely \cite{molchanov2019importance} and how to remove them effectively \cite{suau2018principal,lin2018holistic}. 

\begin{figure}[t]
	\centering
	\includegraphics[width=15cm,]{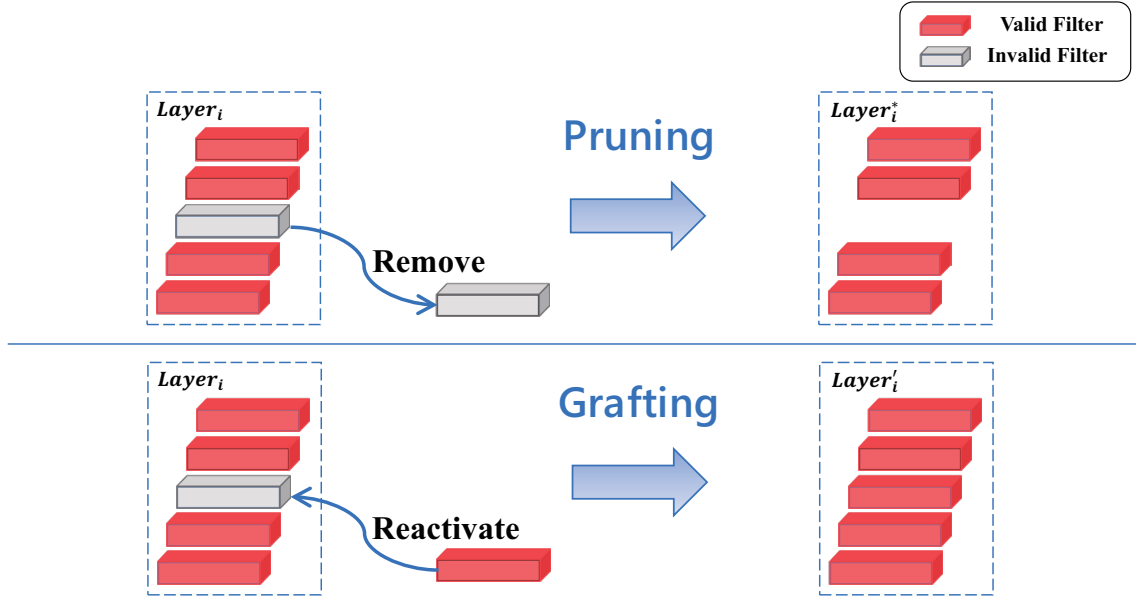}
	\caption{An illustration of the difference between filter pruning and filter grafting. For filter grafting, we graft external information into invalid filters without changing the model structure. (best viewed in color)}
	\label{figure: pruning&grafting}
\end{figure}

However, it is unclear whether abandoning such filters and components directly is the only choice.
What if such traditional \emph{invalid} filters can be turned into \emph{valid} and help the final prediction?
Similar story happens in the ensemble learning like boosting, where a weak classifier is boosted by combining it with other weak classifiers.
In this paper, we investigate the possibility of reactivating invalid filters within the network by bringing outside information.
This is achieved by proposing a novel filter grafting scheme, as illustrated in Figure \ref{figure: pruning&grafting}.
Filter grafting differs from filter pruning in the sense that we reactivate filters by assigning them with new weights from outside, while maintaining the number of layers and filters within each layer as the same. The grafted network has a higher representation capability since more filters in the network are involved to process the information flow. 

A key step in filter grafting is choosing proper information source, \emph{i.e.}, where should we graft the information from. In this paper, we thoroughly study this question and claim that we should graft the information from outside (other networks) rather than inside (within the network). 
We train several networks in parallel. During training at certain epoch, we graft meaningful filters of a network into invalid filters of another network. By performing grafting, each network could learn external information from other networks, leading to a better representation capability \cite{meng2020filter}. Meanwhile, since grafting is operated reciprocally on all networks involved, we find that grafting may lose the information of valid filters when improving invalid filters. To overcome this, we experiment grafting with several cultivation ways and surprisingly find that the impact of grafting and distillation are complementary: distillation mostly enhances valid filters while grafting mostly reactivates invalid filters. Thus we compensate grafting with distillation to gain a universal improvement.
There are four main contributions of this paper:

\begin{itemize}
	\item We propose a new learning paradigm called filter grafting for DNNs. Grafting could reactivate invalid filters to improve the potential of DNNs without changing the network structure.
	
	\item An entropy based criterion and an adaptive weighting strategy are developed to further improve the performance of filter grafting.
	\item We analyze the difference between the distillation and the grafting in terms of their effects on network filters, and find that grafting mostly improves the knowledge of invalid filters while distillation mostly improves that of valid filters. This observation inspires us to compensate grafting with distillation to further increase network performance.
	\item We perform extensive experiments on classification and recognition tasks and show that our method could substantially improve the performance of DNNs. 
\end{itemize}

\section{Related Work}

\begin{table*}[!t]
	\caption{The difference between filter grafting and other learning methods}
	\label{Related_Work_Difference}
	\begin{center}
		\footnotesize
		\begin{tabular}{|c|c|c|c|}
			\hline
			methods& without changing model structure ? &one stage ? & without supervision ?   \\ 
			\hline
			filter pruning \cite{li2016pruning}  & $\times$ &$ \times$ & $\checkmark$
			\\
			\hline
			distillation \cite{hinton2015distilling}&$\checkmark$ &$\times$ &$\times$
			\\	
			\hline	
			deep mutual learning \cite{zhang2018deep}&$\checkmark$ &$\checkmark$ &$\times$
			\\	
			\hline	
			RePr \cite{prakash2019repr} &$\checkmark$ &$\times$ &$\checkmark$
			\\
			\hline
			\textbf{filter grafting} &$\checkmark$ &$\checkmark$ &$\checkmark$
			\\
			\hline
		\end{tabular}
	\end{center}
\end{table*}

\textbf{Filter Pruning.} 
Filter pruning \cite{han2015learning, han2015deep, guo2016dynamic, liu2017learning, lebedev2016fast, wen2016learning} aims to remove the invalid filters to accelerate the inference of the network. \cite{li2016pruning} is the first work that utilizes $l_{1}$ norm criterion to prune unimportant filters. Since then, more criterions came out to measure the importance of the filters. \cite{zhuo2018scsp} utilizes spectral clustering to decide which filter needs to be removed. \cite{suau2018principal} proposes an inherently data-driven method that utilizes Principal Component Analysis (PCA) to specify the proportion of the energy that should be preserved. \cite{wang2018exploring} applies subspace clustering to feature maps to eliminate the redundancy in convolutional filters. Instead of abandoning the invalid filters, filter grafting intends to reactivate them. It is worth noting that even though the motivation of filter grafting is opposite to the pruning, it still involves choosing a proper criterion to decide which filters are unimportant. Thus different criterions used in the pruning are readily applied to the grafting.

\textbf{Distillation and Mutual Learning.} Grafting may involve training multiple networks in parallel. Thus this process is similar to knowledge distillation \cite{hinton2015distilling} and mutual learning \cite{zhang2018deep}. 
Knowledge distillation can be traced back to model compression \cite{bucilua2006model}, where authors demonstrate that the knowledge acquired by a large ensemble of models can be transferred to a single small model. Hinton et al. \cite{hinton2015distilling} generalize this idea to deep neural networks and show a small, shallow network can be improved through a teacher-student framework. Meanwhile, past works have also been focusing on improving the quality and applicability of knowledge distillation and understanding how knowledge distillation works. The work \cite{tang2016recurrent} employs the knowledge transfer learning approach to train RNN using a deep neural network model as the teacher. This is different from the original knowledge distillation, since the teacher used is weaker than the student. 
Lopes et al. \cite{lopes2017data} present a method for data-free knowledge distillation, which is able to compress deep neural networks trained on large-scale datasets to a fraction of their size leveraging only some extra metadata with a pre-trained model. 
The work \cite{mirzadeh2019improved,cho2019efficacy} shows that the student network performance degrades when the gap between student and teacher is large. They introduce multi-step knowledge distillation which employs an intermediate-size network to solve this. 
The main difference between filter grafting and knowledge distillation is that distillation is a `two-stage' process. First, we need to train a large model (teacher), then we use the trained model to teach a small model (student). Meanwhile, grafting is a `one-stage' process, we graft the weight during the training process. However, we demonstrate in the paper that distillation 
and grafting are complementary on the filter level.

The difference between filter grafting and mutual learning is that mutual learning needs a mutual loss to supervise each network to learn and do not generalize well to multiple networks. While grafting does not need supervised loss and performs much better when we add more networks into the training process. Also, we graft the weight at each epoch instead of each iteration, which greatly reduce the communication cost among networks.

\textbf{RePr.} RePr \cite{prakash2019repr} is similar to our work that considers improving the network on the filter level. However, the motivation of RePr is that there exists unnecessary overlaps in the features captured by network filters. RePr first prunes overlapped filters to train the sub-network, then restores the pruned filters and re-trains the full network. In this sense, RePr is a multi-stage training algorithm. In contrast, the motivation of filter grafting is that the filter whose $l_{1}$ norm is smaller contributes less to the network output. Thus the filters that each method operates are different. Also grafting is a one-stage training algorithm which is more efficient. 

\textbf{Grafting.} Grafting is firstly adopted as an opposite operation against pruning in decision tree \cite{webb1997decision}, which adds new branches to the existing tree to increase the predictive performance. \cite{li2012network} graft additional nodes onto the hidden layer of trained neural network for domain adaption. NGA \cite{hu2020exploiting} is proposed to graft front end network onto a trained network to replace its counterpart, which adapts the model to another domain. While in our approach, filters from different networks are weighted and summed to reactivate invalid filters. 

To better illustrate how grafting differs from the above learning types. We draw a table in Table \ref{Related_Work_Difference}. From Table \ref{Related_Work_Difference}, filter grafting is a one stage learning method, without changing network structure and does not need supervised loss.

\section{The Proposed Approach}\label{sec_3}

This section arranges as follows: In Section \ref{sec_3_1}, we study the information source we need during the grafting process; In Section \ref{sec_3_2}, we propose two criterions to calculate the information of filters; In Section \ref{sec_3_3}, we discuss how to effectively use the information for grafting; In Section \ref{sec_3_4}, we extend grafting method to multiple networks and propose entropy-based grafting algorithm. In section \ref{sec_3_5}, we propose grafting+ to overcome the drawbacks of grafting.

\subsection{Information Source for Grafting}\label{sec_3_1}
In the remaining, we call the original invalid filters as 'rootstocks' and call the meaningful filters or information to be grafted as 'scions', which is consistent with botany interpretation for grafting. Filter grafting aims to transfer information (weights) from scions to rootstocks, thus selecting useful information is essential for grafting. In this paper, we propose three ways to get scions.

\subsubsection{Noise as Scions}\label{sec_3_1_1}

A simple way is to graft gaussian noise $ \mathcal{N}(0,\sigma_{t})$ into invalid filters, since Gaussian noise is commonly used for weight initialization of DNNs \cite{kumar2017weight,he2015delving}. 
Before grafting, the invalid filters have smaller $l_{1}$ norm and have little effects for the output.
But after grafting, the invalid filters have larger $l_{1}$ norm and begin to make more effects on the output. 

\begin{equation}\label{noise_decrease}
	\sigma_{t}=a^t(0<a<1)
\end{equation}

We also let $\sigma_{t}$ decrease over time (see \eqref{noise_decrease}), since too much noise may make the model hard to converge. In Section \ref{sec_4_1}, we show that Gaussian noise does not contain useful information thus the improvement is limited.

\subsubsection{Internal Filters as Scions}\label{sec_3_1_2}

\begin{figure}[!h]
	\centering
	\includegraphics[width=15cm]{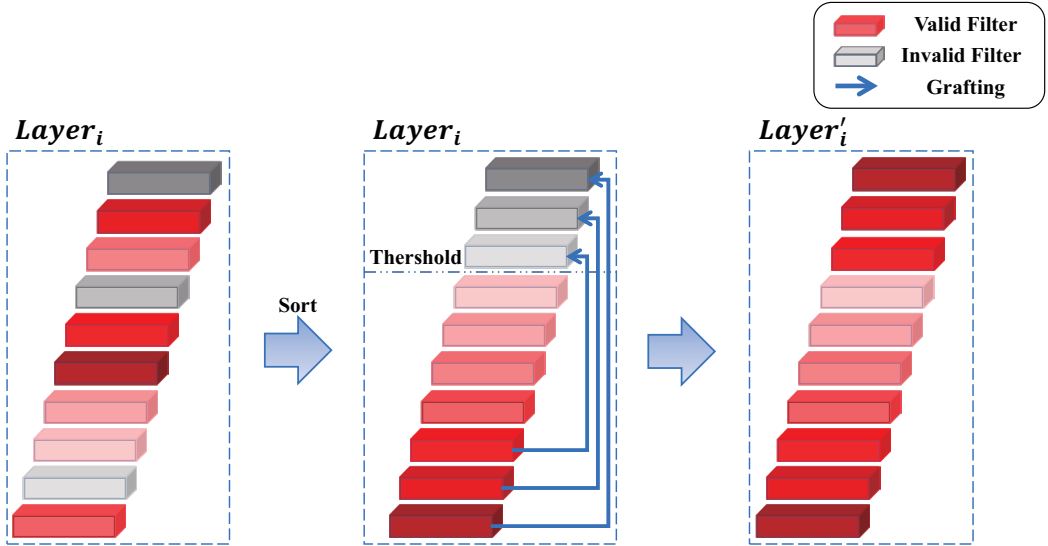}
	\caption{Grafting internal filters. We first sort the filters by $l_{1}$ norm, then graft the weights from filters with larger $l_{1}$ norm into filters with smaller $l_{1}$ norm (best viewed in color).}
	\label{figure: inside_grafting}
\end{figure}

We also experiment adding the internal weights of other filters (whose $l_{1}$ norm is bigger) into the invalid filters (whose $l_{1}$ norm is smaller) of the same network. Specifically, for each layer, we sort the filters by $l_{1}$ norm and set a threshold $\gamma$. For filters whose $l_{1}$ norm are smaller than $\gamma$, we treat these filters as invalid ones. Then we graft the weights of the $i$-th largest filter into the $i$-th smallest filter. This procedure is illustrated in Figure \ref{figure: inside_grafting}.

Since the invalid filters have new weights with larger $l_{1}$ norm, they can be activated to have a bigger influence on the output. But this method does not bring new information to the network since the weights are grafted from exactly the same network. We further evaluate it via the language of information theory. To simplify the proving process, we deal with two filters in a certain layer of the network (See Theorem \ref{theo1}). From Theorem \ref{theo1}, selecting internal filters as scions does not bring new information. The experiment in Section \ref{sec_4_1} is also consistent with our analysis.

\begin{theorem}\label{theo1}
	Suppose there are two filters in a certain layer of the network, denoted as random variables $X$ and $Y$. $Z$ is another variable which satisfies $Z = X+Y$, then $H(X,Y) = H(X,Z) = H(Y,Z)$, where $H$ denotes the entropy from information theory.
\end{theorem}

\begin{Proof}
	We first prove $H(Z|X) = H(Y|X)$:
	\begin{small}
		\begin{align*}
			&H(Z|X) \\
			&= -\sum_{x}p(X=x)\sum_{z}p(Z=z|X=x)\log P(Z=z|X=x)\\
			&= -\sum_{x}p(X=x)\sum_{z}p(Y=z-x|X=x) \log P(Y=z-x|X=x)\\
			&= -\sum_{x}p(X=x)\sum_{y}p(Y=y|X=x)\log P(Y=y|X=x)\\
			& = H(Y|X)
		\end{align*}
	\end{small}
	Then, according to the principle of entropy, we have:
	\begin{align*}
		H(X,Y)& = H(X) + H(Y|X) \\
		&= H(X) + H(Z|X)\\
		&= H(X,Z).\\
	\end{align*}
	By the symmetry of  entropy, the other direction also holds. Thus:
	\[
	H(X,Y) = H(X,Z) = H(Y,Z).
	\]
\end{Proof}

\subsubsection{External Filters as Scions}	\label{sec_3_1_3}

In response to the shortcomings of adding random noise and weights inside a single network, we select external filters from other networks as scions.
Specifically, we could train two networks, denoted as $M_{1}$ and $M_{2}$, in parallel. During training at certain epoch, we graft valid filters' weights of $M_{1}$ into invalid filters of $M_{2}$. Besides, compared to the grafting process in Section \ref{sec_3_1_2}, we make two modifications:
\begin{itemize}
	\item The grafting is processed at layer level instead of filter level, which means we graft the weights of all the filters in a certain layer in $M_{1}$ into the same layer in $M_{2}$, \emph{vice versa}. Since two networks are initialized with different weights, the location of invalid filters are statistically different and only grafting information into part of filters in a layer may break layer consistency (see more analyses and experimental results in the supplementary material).  By performing grafting, the invalid filters of two networks can learn mutual information from each other.
	\item When performing grafting, the inherent information and the extoic information are weighted. Specifically, We use $W_{i}^{M_{2}}$ denotes the weights of the $i$-th layer of $M_{2}$, $W_{i}^{M_{2}^{'}}$ denotes the weights of the $i$-th layer of $M_{2}$ after grafting. Then:
	\begin{equation}\label{weighting}
		W_{i}^{M_{2}^{'}} = \alpha W_{i}^{M_{2}} + (1-\alpha) W_{i}^{M_{1}}~~~ (0<\alpha<1).
	\end{equation}
	Suppose $W_{i}^{M_{2}}$ is more informative than $W_{i}^{M_{1}}$, then $\alpha$ should be larger than $0.5$. 
\end{itemize}

\begin{figure}[!h]
	\centering
	\includegraphics[width=15cm,]{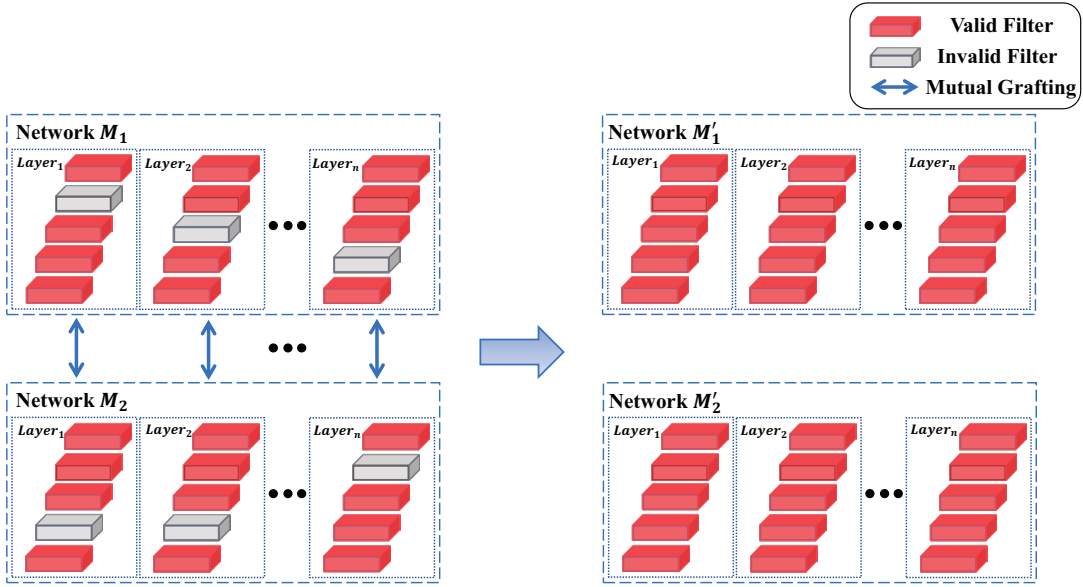}
	\caption{Grafting between two networks. Each network receives the information from the other network (best viewed in color).}
	\label{figure: mutual_grafting}
\end{figure}

The two-networks grafting procedure is illustrated in Figure \ref{figure: mutual_grafting}. There are two key points in grafting: 1) how to calculate the information of $W_{i}^{M_{1}}$ and $W_{i}^{M_{2}}$; 2) how to decide the weighting coefficient $\alpha$.
We thoroughly study these two problems in Section \ref{sec_3_2} and Section \ref{sec_3_3}. Also, we hope to increase the diversity of two networks, thus two networks are initialized differently and some hyper-parameters of two networks are also different from each other (\emph{e.g.}, learning rate, sampling order of data, \emph{etc}).  It is worth noting that when performing grafting algorithm on two networks, the two networks have the same weights after grafting process from \eqref{weighting}. Grafting is only performed at the end of each epoch, and for other iteration steps, since the two networks are learned with different hyper-parameters, their weights are different from each other. Also, this weighting problem disappears when we add more networks ($N>2$) in grafting algorithm. Multiple networks grafting can be found in Section \ref{sec_3_4}.

\subsection{Criterions for Calculating Information of Filters and Layers}\label{sec_3_2}		
In this section, we study two criterions to calculate the information of filters or layers.

\subsubsection{$L_{1}$ norm }\label{sec_3_2_1}	
In previous sections, we use $l_{1}$ norm to measure the information of filters. Denote $W_{i,j}\in \mathbb{R}^{N_{i}\times K\times K }$ as the weight of the $j$-th filter in the $i$-th convolutional layer, where $N_{i}$ is the number of filters in $i$-th layer. Its $l_{1}$ norm can be presented by:
\begin{equation}\label{l1 norm}
	\|W_{i,j}\|_{1}= \sum_{n=1}^{N_{i}}\sum_{k_{1}=1}^{K}\sum_{k_{2}=1}^{K}|W_{i,j}(n,k_{1},k_{2})|.
\end{equation}

The $l_{1}$ norm criterion is commonly used in many researches \cite{li2016pruning,Ye2018Rethinking,Yang2018Soft}. 

\begin{algorithm*}[!t]
	\caption{Entropy-based Multiple Networks Grafting}
	\label{algorithm2}
	\begin{algorithmic}[2]
		\renewcommand{\algorithmicrequire}{\textbf{Input:}}
		\renewcommand{\algorithmicensure}{\textbf{Iteration:}}
		\REQUIRE ~~\\
		Number of networks $K$, $M_{k}$ denotes the $k$-th network; Number of layers $L$; 
		Training iterations $\mathcal N = \{1,\ldots, N_{max}  \}$; Number of iterations for each epoch $N_{T}$; Training dataset $\mathcal D$;
		Initial weights for each layer of each network $\{\mathbf{W}_{l}^{M_{k}}:k=1,\ldots, K; l=1,\ldots, L \}$; Different hyper-parameters for each network $\{\mathbf{\lambda}_k:k=1,\ldots K\}$.	\ENSURE ~~\\
		\textbf{for} $n = 1$ to $N_{max}$   
		\\
		\qquad\textbf{for} $\mathbf{k}\in\{1,\ldots K\}$, $\mathbf{l}\in\{1,\ldots L\}$ \textbf{parallel do}
		\\
		\qquad\qquad Update model parameters $\mathbf{W}^{M_{k}}_{l}$ based on $\mathcal D$ with $\mathbf{\lambda}_k$ ~~~//Update model weights at each iteration.
		\\
		\qquad\qquad \textbf{if} $n~mod~N_{T} = 0$ 
		\\
		\qquad\qquad\qquad Get the weighting coefficient $\alpha$ from \eqref{coefficient} ~~~~~~~~//Graft model weights at each epoch.
		\\
		\qquad\qquad\qquad  $\mathbf{W}^{M_{k}}_{l} = \alpha \mathbf{W}^{M_{k}}_{l} + (1-\alpha)\mathbf{W}^{M_{k-1}}_{l}$  
		\\
		\qquad\qquad \textbf{end if}
		\\
		\qquad\textbf{end for}
		\\
		\textbf{end for}
		\\
	\end{algorithmic}
\end{algorithm*}

\subsubsection{Entropy}	\label{sec_3_2_2}

While $l_{1}$ norm criterion only concentrates on the absolute value of filter's weight, we pay more attention to the variation of the weight. 
We suppose each value of $W_{i,j}$ is sampled from a distribution of a random variable $X$ and use the entropy to measure the distribution. Suppose the distribution satisfies $P(X=a)=1$, then each single value in $W_{i,j}$ is the same and the entropy is 0. 
While calculating the entropy of continuous distribution is hard, we follow the strategy from \cite{shwartz2017opening,cheng2019utilizing}. We first convert continuous distribution to discrete distribution. Specifically, we divide the range of values into $m$ different bins and calculate the probability of each bin. Finally, the entropy of the variable can be calculated as follows:
\begin{equation}\label{binning}
	H(W_{i,j}) =-\sum_{k=1}^{B}p_k\log p_k,
\end{equation}
where $B$ is the number of bins and $p_{k}$ is the probability of bin $k$. A smaller score of $H(W_{i,j})$ means the filter has less variation (information).

Suppose layer $i$ has $C$ filters, then the total information of the layer $i$ is:
\begin{equation}\label{info_layer_pre}
	H(W_{i}) = \sum_{j=1}^{C} H_{i,j}.
\end{equation}
One problem of \eqref{info_layer_pre} is that it neglects the correlations among the filters since \eqref{info_layer_pre} calculates each filter's information independently. To keep the layer consistency, we directly calculate the entropy of the whole layer's weight $W_{i}\in \mathbb{R}^{N_{i}\times N_{i+1}\times K\times K }$ as follows:
\begin{equation}\label{info_layer}
	H(W_{i})=-\sum_{k=1}^{B}p_k\log p_k.
\end{equation}
Different from \eqref{binning}, the values to be binned in \eqref{info_layer} are from the weight of the whole layer instead of a single filter. In the supplementary material, we prove layer consistency is essential for grafting algorithm.

\subsection{Adaptive Weighting in Grafting}\label{sec_3_3}
\begin{figure}[!h]
	\centering
	\includegraphics[width=15cm,]{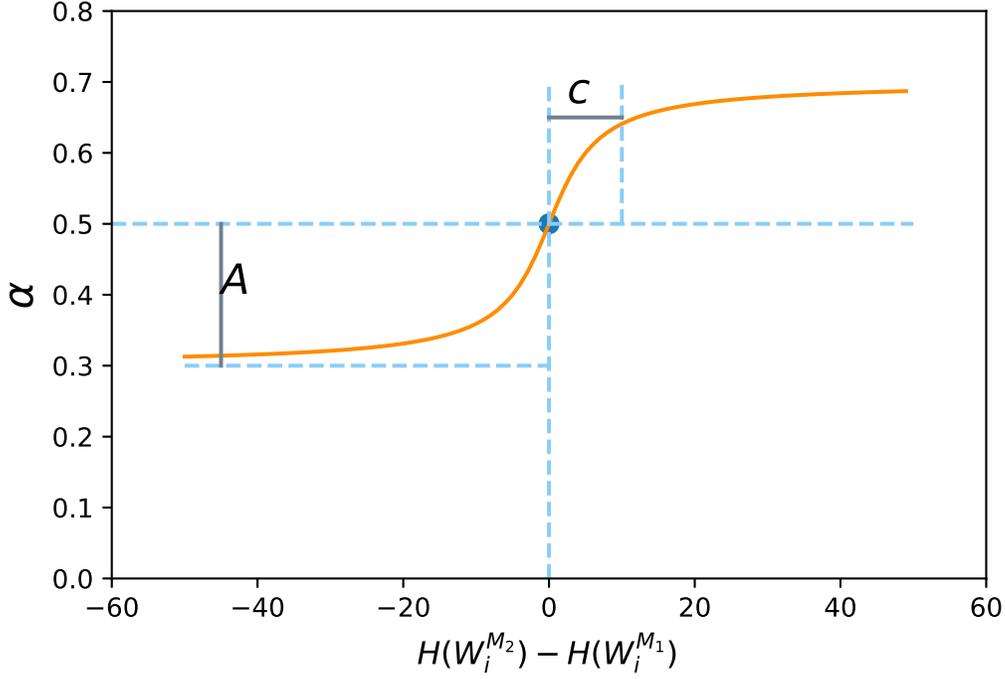}
	\caption{The adaptive coefficient in grafting process.}
	\label{figure: adaptive_w}
\end{figure}

In this part, we propose an adaptive weighting strategy for weighting two models' weight from \eqref{weighting}.  Denote $W_{i}^{M_{1}}$ and $H(W_{i}^{M_{1}})$ as the weight and information of layer $i$  in network $M_{1}$, respectively. The calculation of $H(W_{i}^{M_{1}})$ can be referred to \eqref{info_layer}. We enumerate two conditions that need to be met for calculating the coefficient $\alpha$.
\begin{itemize}
	\item The coefficient $\alpha$ from \eqref{weighting} should be equal to 0.5 if $H(W_{i}^{M_{2}}) = H(W_{i}^{M_{1}})$ and 
	larger than 0.5 if $H(W_{i}^{M_{2}}) > H(W_{i}^{M_{1}})$.
	\item Each network should contain part of self information even though $H(W_{i}^{M_{2}}) \gg H(W_{i}^{M_{1}})$ or $H(W_{i}^{M_{2}}) \ll H(W_{i}^{M_{1}})$.
\end{itemize}
In response to the above requirements, the following adaptive coefficient is designed: 
\begin{equation}\label{coefficient}
	\alpha=A\ast(arctan(c\ast(H(W_{i}^{M_{2}})-H(W_{i}^{M_{1}}))))+0.5,
\end{equation}
where $A$ and $c$ in \eqref{coefficient} are fixed hyper-parameters, and $\alpha$ is the coefficient of \eqref{weighting}. We further depict a picture in 
Figure \ref{figure: adaptive_w}. We can see this function well satisfies the above conditions.

\subsection{Extending Grafting to Multiple Networks}\label{sec_3_4}

\begin{figure}[!h]
	\centering
	\includegraphics[width=15cm,]{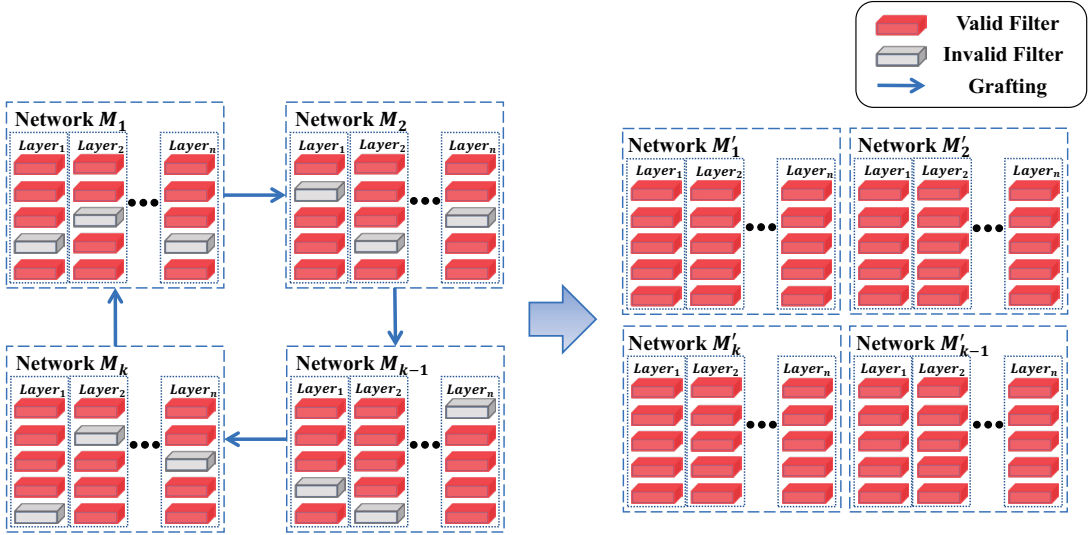}
	\caption{Grafting with multiple networks. The network $M_{k}$ accepts information from $M_{k-1}$. (best viewed in color)}
	\label{figure: multiple_grafting}
\end{figure}

Grafting method can be easily extended to a multi-networks case, as illustrated in Figure \ref{figure: multiple_grafting}. At the end of each epoch during training, each network $M_{k}$ accepts the information from $M_{k-1}$. After certain training epochs, each network contains information from all the other networks. The weighting coefficient is also calculated adaptively. From Section \ref{sec_4_2}, we find that by using grafting to train multiple networks, all networks involved could expect a performance gain. 
We propose our entropy-based grafting in Algorithm \ref{algorithm2}. It is worth noting that grafting is performed on multiple networks in parallel, which means when we use $\mathbf{W}^{M_{k-1}}_{l}$ to update $\mathbf{W}^{M_{k}}_{l}$, $\mathbf{W}^{M_{k-1}}_{l}$ has not been updated by grafting yet.

\subsection{Cultivating the Grafted Network (Grafting+)}\label{sec_3_5}
Grafting aims to improve the quality of invalid filters. However, when performing grafting, the weight of invalid filters may also be grafted to the other networks which degrade the knowledge of valid filters.
The experimental results in Section 5 also verify this assumption. To alleviate this problem, we employ distillation in grafting system since we find that the impact of grafting and distillation are complementary: distillation mostly enhances valid filters while grafting mostly reactivates invalid filters (verified in Section \ref{sec:exp}). 

Suppose there are $C$ students network and $M$ teachers network, besides performing grafting among student networks, we also perform distillation among teachers and students. That is, except for the base cross-entropy loss, we also add distilled loss on each student and the total loss can be expressed as:
\begin{equation}
	L_{s_{c}} = L_{C_{s_{c}}} + L_{KD_{t\rightarrow s_{c}}},
\end{equation}
where $ L_{C_{s_{c}}}$ is the typical cross entropy loss:
\begin{equation}
	L_{C_{s_{c}}} = -\sum_{i=1}^{N}\sum_{k=1}^{K}I(y_{i},k)\log(p_{s_{c}}^{k}(x)).
\end{equation}
The indicator function $I$ is defined as 

\begin{equation}
	I(y_{i},k) = 
	\left\{
	\begin{array}{c}
		1 ~~~~~~ y_{i} = k   \\
		0 ~~~~~~ y_{i}\neq k  \\
	\end{array}
	\right.
\end{equation}

$L_{KD_{t\rightarrow s_{c}}}$ is the knowledge distillation loss defined as:

\begin{equation}\label{eq_distill}
	L_{KD_{t\rightarrow s_{c}}}= - \sum_{i=1}^{N}\tau^{2}\sum_{k=1}^{K}\bar{p}^{t}_{k}(x_{i}) \log p^{s_{c}}_{k}(x_{i}),
\end{equation}

where $\bar{p}^{t}_{k}(x_{i})$ is the average probability of all the teachers given $x_{i}$:

\begin{equation}
	\bar{p}^{t}_{k}(x_{i}) = \frac{1}{M} \sum_{m=1}^{M} p^{t_{m}}_{k}(x_{i}).
\end{equation}

\begin{figure}[!t]
	\centering
	\includegraphics[width=15cm,]{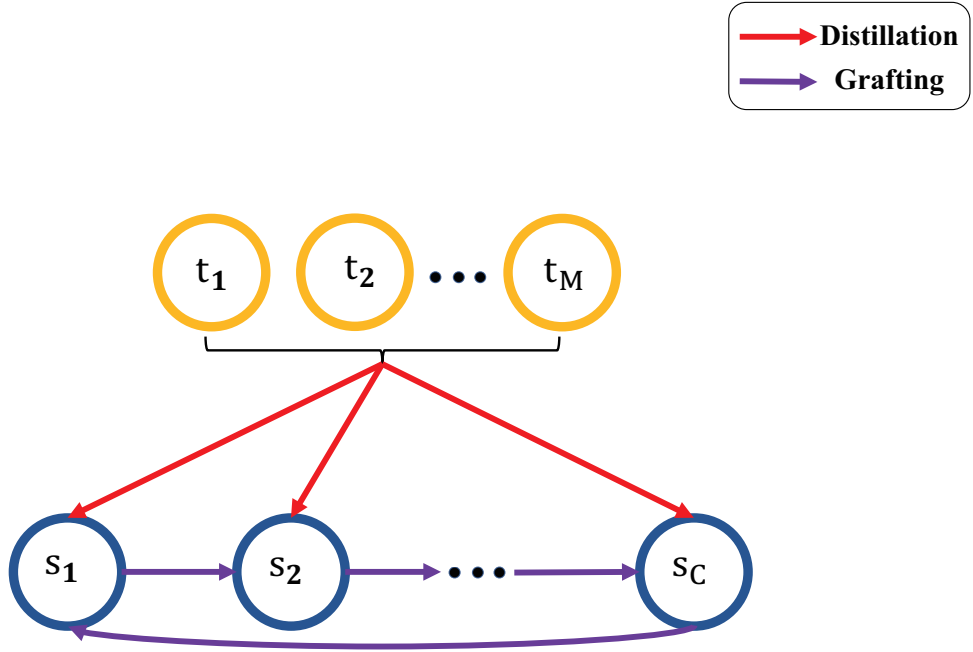}
	\caption{This is the overall framework of grafting+. KD loss is added to the student's cross-entropy loss. (best viewed in color)}
	\label{figure: grafting+}
\end{figure}

The overall framework is presented in Figure \ref{figure: grafting+}. In Section \ref{sec:exp}, we also find that adding more teachers can further
increase the performance. It is worth noticing that even though we bring more networks in the framework, the training can be efficiently implemented parallelly. Also we only need one network for inference thus no additional inference cost is required.

\section{Experiments}
\label{sec:exp}

We conduct our experiments in four aspects:
\begin{itemize}
	\item In Section \ref{sec_4_1}, we examine the basic components of grafting, \emph{i.e.}, the information source and the information criterion. Through the experiments, we show that external filters as scions achieves the best result. Also the entropy-based criterion is better than $l_{1}$ norm-based criterion. 
	\item In Section \ref{sec_4_2}, we perform extensive experiments to verify the effectiveness of grafting and show grafting does improve the invalid filters, leading to a network with better representation ability.
	\item In Section \ref{sec_4_3}, we find that grafting may affect the information of valid filters. However, we observe that the impact of grafting and distillation are complementary: distillation mostly enhances valid filters while grafting mostly reactivates invalid filters. Thus we combines grafting with distillation as a way of cultivation to gain a universal improvement.
	\item In Section \ref{sec_4_4}, we evaluate the performance of grafting together with cultivation and prove that the integral grafting process could universally improve the filters. 
\end{itemize}

\subsection{Information Sources and Criterion}\label{sec_4_1}

\textbf{Selecting Useful Information Source}: We propose three ways to get scions in Section \ref{sec_3} and experimentally examine the three ways on CIFAR-10 and CIFAR-100 datasets \cite{cifar10} in Table \ref{table:branches}. Vanilla DNN training without grafting is taken as the baseline. All the methods use MobileNetV2 as the base model. For a fair comparison, the same hyper-parameters are deployed for each method: mini-batch size (256), optimizer (SGD), initial learning rate (0.1), momentum (0.9), weight decay (0.0005), number of epochs (200), learning rate decay (0.1 at every 60 epochs).
'External' here involves training two networks in parallel. In practice, we find the performance of each network in the 'external' method is very close to each other. Thus in the remaining, we always record the first network's performance.

\begin{table}[!h]
	\caption{Comparison of different scion sources. }
	\begin{center}
		\begin{tabular}{|c|c|c|} 
			\hline 
			&CIFAR-10 &CIFAR-100\\ 
			\hline
			baseline&92.42&71.44\\
			noise&92.51&72.34\\
			internal&92.68&72.38\\
			external&\textbf{92.94}&\textbf{72.90}\\
			\hline
		\end{tabular}\\
	\end{center}
	\label{table:4_1}
\end{table}

From Table \ref{table:4_1}, the performance of `internal scions' is similar to 'noise', since we prove in Theorem \ref{theo1} that choosing internal filters as scions does not bring new information to the network. While choosing external filters as scions achieves the best result among the three methods. In the remaining, all the grafting experiments choose external filters as scions.

\textbf{Comparison of $L_{1}$ norm \& Entropy Criterions}: We propose two criterions to measure the inherent information of filters in Section \ref{sec_3_2}. In this part, we quantitatively evaluate the $l_{1}$ norm-based grafting and the entropy-based grafting on CIFAR-10 and CIFAR-100 dataset. The results are listed in Table \ref{table:norm_entropy}. Two networks are used for grafting, with an identical model structure and training hyper-parameters. From Table \ref{table:norm_entropy}, we can find that, entropy-based grafting beats $l_{1}$ norm-based grafting on every model and dataset setting.

\begin{table}[!h]
	\caption{Comparison of grafting by $l_{1}$ norm \& entropy. }
	\begin{center}
		\begin{tabular}{|c|c|c|c|} 
			\hline 
			model&method&CIFAR-10 &CIFAR-100\\ 
			\hline
			&baseline&92.83&69.82\\
			ResNet32&$l_{1}$ norm&93.24&70.69\\
			&entropy&\textbf{93.33}&\textbf{71.16}\\
			\hline
			&baseline&93.50&71.55\\
			ResNet56&$l_{1}$ norm&94.09&72.73\\
			&entropy&\textbf{94.28}&\textbf{73.09}\\
			\hline
			&baseline&93.81&73.21\\
			ResNet110&$l_{1}$ norm&94.37&73.65\\
			&entropy&\textbf{94.60}&\textbf{74.70}\\
			\hline
			&baseline&92.42&71.44\\
			MobileNetV2&$l_{1}$ norm&92.94&72.90\\
			&entropy&\textbf{93.53}&\textbf{73.26}\\
			\hline
		\end{tabular}
	\end{center}
	\label{table:norm_entropy}
\end{table}

In the rest of the experiments, when performing grafting, we always use external filters as scions and apply entropy based criterion.

\begin{table*}[!h]
	\begin{center}
		\begin{tabular}{c|c|c|c|c|c|c} 
			\hline 
			Dataset&method&ResNet32&ResNet56&ResNet110&MobileNetV2&WRN28-10\\
			\hline 
			\multirow{4}{*}{CIFAR-10}
			&baseline&92.83&93.50&93.81&92.42&95.75\\
			&mutual learning \cite{zhang2018deep}&92.80&--&--&--&95.66\\
			&RePr \cite{prakash2019repr}&93.90&--&94.60&--&--\\
			&filter grafting&\textbf{93.94}&\textbf{94.73}&\textbf{94.96}&\textbf{94.20}&\textbf{96.40}\\
			\hline 
			\multirow{4}{*}{CIFAR-100}
			&baseline&69.82&71.55&73.21&71.44&80.65\\
			&mutual learning \cite{zhang2018deep}&70.19&--&--&--&80.28\\
			&RePr \cite{prakash2019repr}&69.90&--&73.60&--&--\\
			&filter grafting&\textbf{71.28}&\textbf{72.83}&\textbf{75.27}&\textbf{74.15}&\textbf{81.62}\\
			\hline 
		\end{tabular}
	\end{center}
	\caption{Comparion of filter grafting with other learning methods. `--' denotes the result is not reported in the corresponding paper.}
	\label{table:4_3}
\end{table*}

\subsection{Evaluating Grafting on Different Datasets}\label{sec_4_2}

\textbf{CIFAR-10 and CIFAR-100}: We first compare grafting with other methods on CIFAR-10 and CIFAR-100 dataset. The results are shown in Table \ref{table:4_3}. For a fair comparison, `mutual learning' and `filter grafting' all involve training two networks. The networks in mutual learning and grafting all have the same network structures. The difference between mutual learning and grafting is that mutual learning trains two networks with another strong supervised loss and communication costs are heavy between networks. One should carefully choose the coefficient for mutual supervised loss and main loss when using the mutual learning method. While for grafting, transferring weights does not need supervision. We graft the weights by utilizing entropy to adaptively calculate the weighting coefficient which is more efficient. The results from Table \ref{table:4_3} show that filter grafting achieves the best results among all the learning methods. We also examine the effect of multi-networks grafting in Table \ref{table:4_4}.

\begin{table}[!h]
	\caption{Grafting with multiple networks (MobileNetV2). }
	\begin{center}
		\begin{tabular}{|c|c|c|} 
			\hline 
			method&CIFAR-10 &CIFAR-100\\ 
			\hline
			baseline& 92.42 & 71.44 \\
			\hline
			2 models grafting&94.20&74.15\\
			3 models grafting&94.55&76.21\\
			4 models grafting&95.23&77.08\\
			6 models grafting&\textbf{95.33}&\textbf{78.32}\\
			8 models grafting&95.20&77.76\\
			\hline
			6 models ensemble&94.09&76.75\\
			\hline
		\end{tabular}\\
	\end{center}
	\label{table:4_4}
\end{table}

As we raise the number of networks, the performance gets better. For example, the performance with 6 models grafting could outperform the baseline by about 7 percent which is a big improvement. The reason is that MobileNetV2 is based on depth separable convolutions, thus the filters may learn insufficient
knowledges. Filter grafting could help filters learn complementary knowledges from other networks, which greatly improves the network's potential. Also it is worth noting that the result of 6 models grafting is even better than 6 models ensembles. But unlike ensemble, grafting only maintains one network for testing. However, the performance stagnates when we add the number of models to 8 in grafting algorithm. We assume the cause might be that the network receives too much information from outside which may affect its self-information for learning. How to well explain this phenomenon is an interesting future work.

\textbf{ImageNet}: To test the performance of grafting on a larger dataset, we also validate grafting on ImageNet \cite{deng2009imagenet}, an image classification dataset with over 14 million images. We compare grafting with the baseline on ResNet18, ResNet34 and ResNet50 models. The baseline hyper-parameters' setting is consistent with official PyTorch setting for ImageNet\footnote{https://github.com/pytorch/examples/tree/master/imagenet}: minibatch size (256), initial learning rate (0.1), learning rate decay (0.1 at every 30 epochs), momentum (0.9), weight decay (0.0001), number of epochs (90) and optimizer (SGD). To increase the training diversity, we use different learning rates and data loaders for two networks when performing grafting. The other hyper-parameters' setting is consistent with the baseline. The results inTable \ref{table:Imagenet} shows that grafting  can also handle larger datasets.

\begin{table}[!h]
	\caption{Grafting on ImageNet Dataset}
	\begin{center}
		\begin{tabular}{|c|c|c|c|} 
			\hline 
			model&method&top-1&top-5\\ 
			\hline
			\multirow{2}{*}{ResNet-18}&baseline&69.15&88.87\\
			&grafting&\textbf{71.19}&\textbf{90.01}\\
			\hline
			\multirow{2}{*}{ResNet-34}&baseline&72.60&90.91\\
			&grafting&\textbf{74.58}&\textbf{92.05}\\
			\hline
			\multirow{2}{*}{ResNet-50}&baseline&75.92&92.81\\
			&grafting&\textbf{76.76}&\textbf{93.34}\\
			\hline
		\end{tabular}\\
	\end{center}
	\label{table:Imagenet}
\end{table}

\textbf{Market1501 and Duke}: Grafting is a general training method for convolutional neural networks. Thus grafting can not only apply to the classification task but also other computer vision tasks. In this part, we evaluate the grafting on Person re-identification (ReID) task \cite{deng2018image, wei2018person, lin2018multi, wang2018transferable}, an open set retrieval problem in distributed multi-camera surveillance, aiming to match people appearing in different non-overlapping camera views. We conduct experiments on two person ReID datasets: Market1501 \cite{market1505} and DukeMTMC-ReID (Duke) \cite{Ristani2016Performance,zheng2017unlabeled}.
The baseline hyper-parameters' setting is consistent with \cite{zhou2019osnet}: mini-batch size (32), pretrained (True), initial learning rate (0.1), learning rate decay (0.1 at every 20 epochs), number of epochs (60). Besides data loaders and learning rate, the other hyper-parameters' setting is consistent with the baseline. Table \ref{table:ReID} shows that for each model and each dataset, grafting performs better than the baseline. Besides, as mentioned before, increasing the number of networks in grafting can further improve the performance.

\begin{table}[!h]
	\caption{Grafting on ReID Task}
	\begin{center}
		\small
		\begin{tabular}{|c|c|cc|cc|} 
			\hline 
			model &method&\multicolumn{2}{c|}{Market1501} &\multicolumn{2}{c|}{Duke}\\ 
			&&mAP&rank1&mAP&rank1\\ 
			\hline
			&baseline&67.6&86.7&56.2&76.2\\
			ResNet50&2 models&70.6&87.8&60.8&79.8\\
			&4 models&\textbf{73.33}&\textbf{89.2}&\textbf{62.1}&\textbf{79.8}\\
			\hline
			&baseline&56.8&81.3&47.6&71.7\\
			MobileNetV2&2 models&63.7&85.2&53.4&76.1\\
			&4 models&\textbf{64.5}&\textbf{85.8}&\textbf{54.3}&\textbf{76.3}\\
			\hline
		\end{tabular}\\
	\end{center}
	\label{table:ReID}
\end{table}

\begin{figure}[!h]
	\centering
	\includegraphics[width=15cm,]{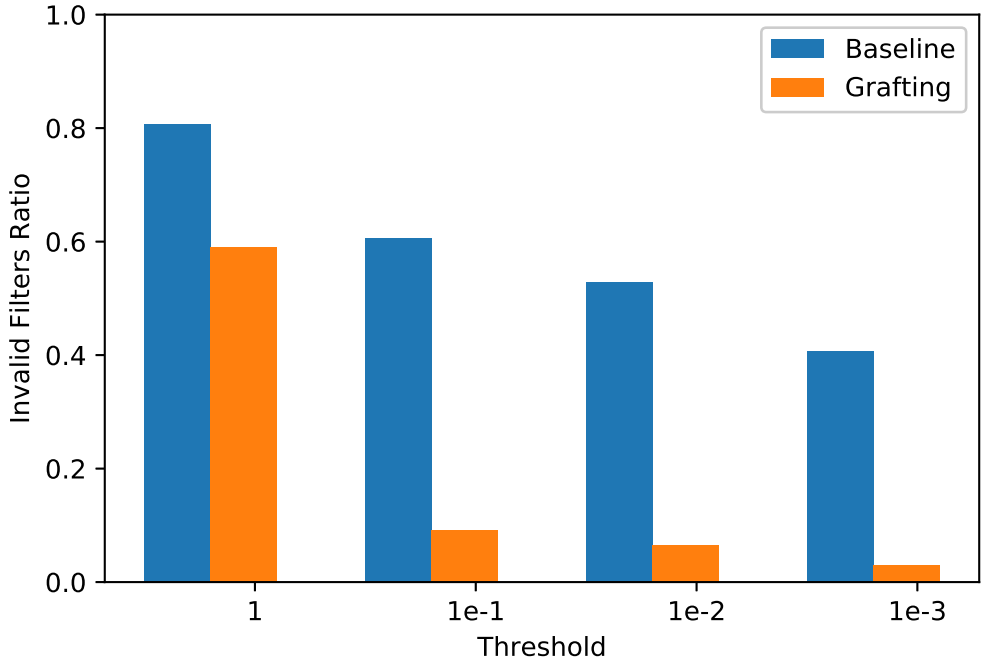}
	\caption{Ratio of filters whose $l_{1}$ norm under some threshold. }
	\label{figure: l1_threshold}
\end{figure}

\begin{figure}[!h]
	\centering
	\includegraphics[width=15cm,]{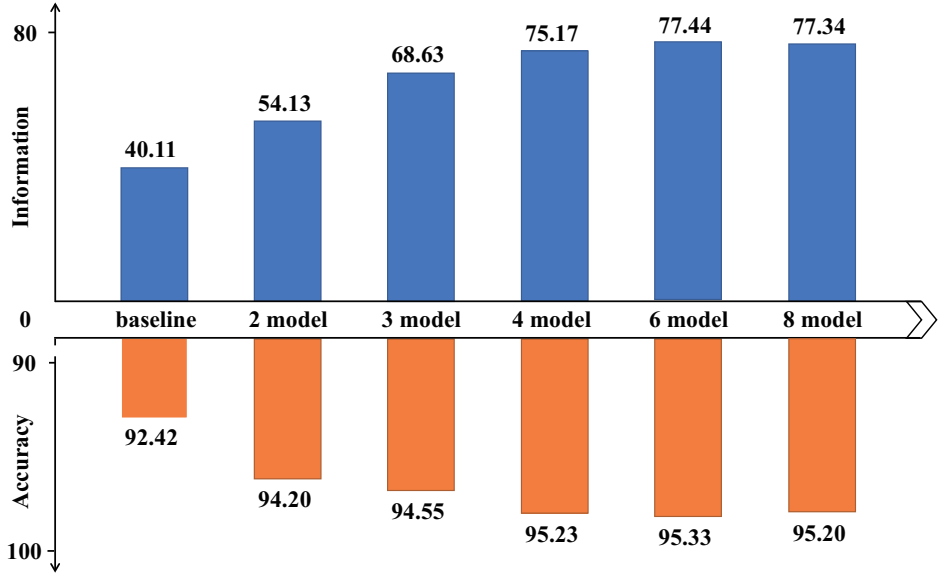}
	\caption{ Entropy and accuracy of the baseline network and grafted network. The network's information is defined as the sum of all the layers' entropy in a \textbf{single} network. The $x$ axis denotes the number of networks parallelly trained in grafting algorithm. }
	\label{figure: entropy_compare}
\end{figure}

\textbf{Effectiveness of Grafting}: In this part, we further analyze the effectiveness of the grafting method. To prove grafting does improve the potential of the network, we calculate the number of invalid filters and information gain after the training process. We select MobileNetV2, which is trained on CIFAR-10 with grafting algorithm, for this experiment. The same network structure without grafting is chosen as the baseline. Experimental results are reported in Figure \ref{figure: l1_threshold} and Figure \ref{figure: entropy_compare}. 

From Figure \ref{figure: l1_threshold}, under the threshold of 1e-3, there are about 50\% filters are invalid or unimportant for the base network, whereas the grafted network only has a small part of filters counted as `invalid', which shows grafting does help network reduce invalid filters.
From Figure \ref{figure: entropy_compare}, the model trained by grafting contains more information than the baseline. Also, the network can gain more information by training multiple networks for grafting method.  Thus from the above analysis, we confirm that grafting could improve the potential of neural networks. 
More analyses can be found in the supplementary material, including the evaluation of invalid filters' locations, necessity of keeping layer consistency and efficiency of adaptive weighting strategy.

\begin{table*}[!h]
	\caption{This table records the density of valid filters and invalid filters for each method.  } 
	\small
	\begin{center}
		\begin{tabular}{|c|c|c|c|c|c|c|c|} 
			\hline 
			Dataset&Threshold&\makecell[tc]{Valid filters\\number}&\makecell[tc]{Invalid filters\\number}&Method &\makecell[tc]{
				Average $l_{1}$ norm \\ of valid filters}& \makecell[tc]{Average $l_{1}$ norm of\\ invalid filters}&\makecell[tc]{Test set\\accuracy} \\ 
			\hline
			\multirow{6}{*}{CIFAR-10}&
			\multirow{3}{*}{0.1}&\multirow{3}{*}{1936}&\multirow{3}{*}{96}
			&baseline&6.982&0.003&93.64\\
			&&&&distillation&\textbf{13.648}&0.000&93.70\\
			&&&&grafting&6.427&\textbf{1.633}&\textbf{94.14}\\
			\cline{2-8}
			&\multirow{3}{*}{1}&\multirow{3}{*}{1920}&\multirow{3}{*}{112}
			&baseline&7.035&0.095&93.64\\
			&&&&distillation&\textbf{13.762}&0.000&93.70\\
			&&&&grafting&6.460&\textbf{1.757}&\textbf{94.14}\\
			\hline
			\multirow{6}{*}{CIFAR-100}&
			\multirow{3}{*}{0.1}&\multirow{3}{*}{1968}&\multirow{3}{*}{64}
			&baseline&12.086&0.001&71.26\\
			&&&&distillation&\textbf{26.914}&0.000&71.92\\
			&&&&grafting&10.802&\textbf{1.765}&\textbf{72.60}\\
			\cline{2-8}
			&\multirow{3}{*}{1}&\multirow{3}{*}{1961}&\multirow{3}{*}{71}
			&baseline&12.126&0.075&71.26\\
			&&&&distillation&\textbf{27.011}&0.000&71.92\\
			&&&&grafting&10.830&\textbf{1.879}&\textbf{72.60}\\
			\hline
		\end{tabular}\\
	\end{center}
	\label{table:branches}
\end{table*}

\subsection{Cultivating with Distillation (Grafting+)}\label{sec_4_3}

From grafting procedure, when we graft multiple networks, even though the information of invalid filters can be replenished by valid filters of other networks, the valid filters may also be affected by invalid filters. It is not clear whether the lost information of valid filters can be recovered during training.
Thus in this experiment, we evaluate how distillation and grafting influence the valid and invalid filter. Different thresholds are set to determine which filters are valid or invalid. Specifically, all the filters are ranked according to their $l_{1}$ norm values. For filters whose $l_{1}$ norms are larger than the threshold, we consider these filters as valid. Inversely, filters are considered to be invalid if their $l_{1}$ norms are smaller than the threshold. Then for each network trained by different methods, we calculate its average $l_{1}$ norm of both the valid and the invalid filter. Datasets and training setting are listed below:

\textbf{Training datasets:} The CIFAR-10 and CIFAR-100 datasets are selected for this experiment. The datasets consist of 32 $\times$ 32 color images with objects from 10 and 100 classes respectively. Both are split into a 50000 images train set and a 10000 images test set. The Top-1 accuracy is used as the evaluation metric.

\textbf{Training setting:}  The baseline training settings are listed as follows: mini-batch size (256), optimizer (SGD), initial learning rate (0.1), momentum (0.9), weight decay (0.0005), number of epochs (200) and learning rate decay (0.1 at every 60 epochs). Standard data augmentation is applied to the dataset. For knowledge distillation hyper-parameters, we set $\tau = 2$ in (\ref{eq_distill}), which are consistent with the work \cite{yang2019snapshot}. For the hyper-parameters regarding filter grafting, we use the same setting from \cite{meng2020filter}, where grafting is performed at the end of each epoch with $A = 0.4$ and $d=500$ in (\ref{coefficient}). We use ResNet-56 as the baseline model. Filter grafting involves two ResNet-56 networks that learn knowledge from each other and one network is selected for testing. Knowledge distillation uses ResNet-110 as the teacher and ResNet-56 as the student. After training, the student network is used for testing.

\begin{figure}[!h]
	\centering
	\includegraphics[width=15cm,]{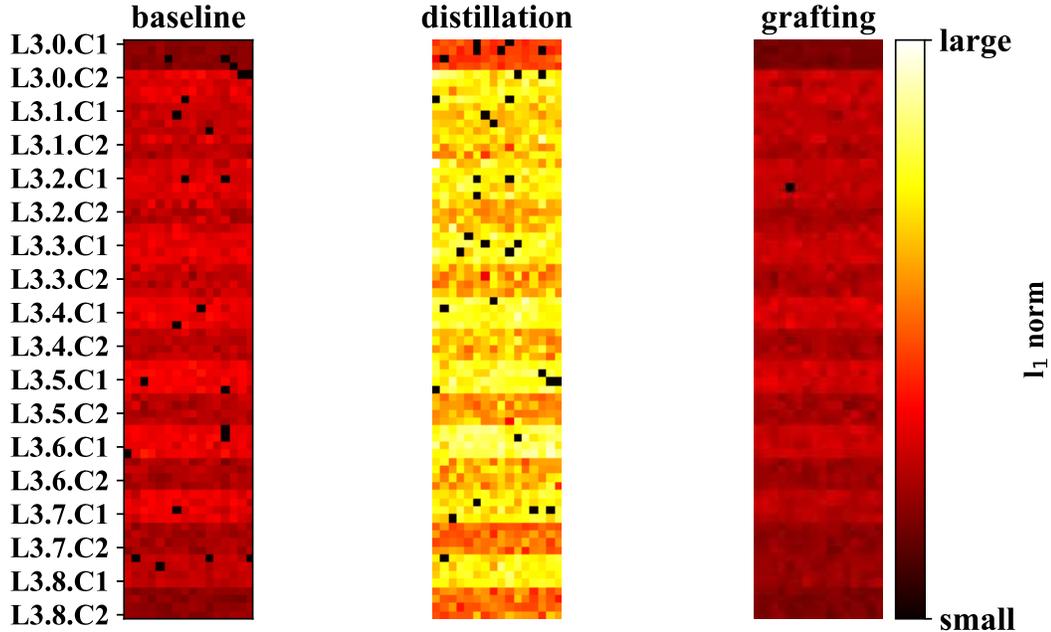}
	\caption{This figure shows the filter weight of each layer. Each filter is represented by a colored cube which is relational to the value of its $l_{1}$ norm. The black cubes represent the filters whose $l_{1}$ norm are very close to 0.  (best viewed in color)}
	\label{figure:bgd}
\end{figure}

It is worth noting that we identify a filter as valid or invalid by the baseline model with a threshold. To see how grafting and distillation influence filters, the number of valid and invalid filters is fixed for baseline, grafting and distillation. Then we calculate the average $l_{1}$ norm of such filters. The results are listed in Table \ref{table:branches}. We can see that both knowledge distillation and filter grafting improve the accuracy of the baseline model. However, they behave differently in the filter level. For knowledge distillation, it greatly densifies the knowledge of valid filters and sparsifies invalid filters. For example, the $l_{1}$ norms of invalid filters trained by distillation is very close to 0. On the other hand, filter grafting mostly densifies invalid filters since the $l_{1}$ norms of invalid filters are prominently improved by the grafting algorithm. We further visualize the filters of the networks trained by each method in Figure \ref{figure:bgd}. We can see that distillation could globally densify the knowledge of the network. However, the network trained by distillation has more invalid filters ($l_{1}$ norm close to 0) than baseline. In contrast, grafting could greatly densify the invalid filters and keep more filters functional in networks.  This observation means that \emph{the two methods boost the neural network in an opposite way which is naturally complementary.} So it inspires us to design a unified framework to further enhance DNNs. We term the new framework as `grafting+', a integral grafting process which is shown in Figure \ref{figure: grafting+}.

\begin{table*}[!t]
	\caption{This table records the accuracy of CIFAR-100 for each method. Grafting trains two student networks simultaneously while distillation maintains one teacher network and one student network. Grafting+ involves one teacher network and two students networks. For each method, we use the single student network for testing. }
	\begin{center}
		\begin{tabular}{|c|c|c|c|} 
			\hline 
			Teacher&Student & Method&CIFAR-100 \\ 
			\hline
			\multirow{2}{*}{---}&	\multirow{4}{*}{ResNet-32}
			&baseline & 69.87 $\pm$ 0.42\\
			&&filter grafting&70.76$\pm$ 0.33\\
			\cline{1-1} 
			\multirow{2}{*}{ResNet-56}
			&&knowledge distillation&72.31$\pm$ 0.23\\
			&&GD&\textbf{72.69$\pm$ 0.26}\\
			\hline
			
			\hline
			\multirow{2}{*}{---}&	\multirow{4}{*}{ResNet-56}
			&baseline&71.87$\pm$ 0.42\\
			&&filter grafting&72.00$\pm$ 0.03\\
			\cline{1-1} 
			\multirow{2}{*}{ResNet-110}
			&&knowledge distillation&73.42$\pm$ 0.29\\
			&&GD&\textbf{74.28$\pm$ 0.19}\\
			\hline
			
			\hline
			\multirow{2}{*}{---}&	\multirow{4}{*}{MobileNetV2}
			&baseline&72.4$\pm$ 0.32\\
			&&filter grafting&73.79$\pm$ 0.18\\
			\cline{1-1} 
			\multirow{2}{*}{ResNet-110}
			&&knowledge distillation&74.98$\pm$ 0.14\\
			&&DGD&\textbf{75.55$\pm$ 0.03}\\
			\hline
		\end{tabular}\\
	\end{center}
	\label{table:grafting+_cifar}
\end{table*}

\begin{table*}[!t]
	\caption{This table records the average $l_{1}$ norm of valid filters and invalid filters for each method. The backbone is ResNet-56. }
	\begin{center}
		\begin{tabular}{|c|c|c|c|c|c|} 
			\hline 
			Dataset&Threshold&Method &\makecell[tc]{Average $l_{1}$ norm 
				of \\ valid filters}& \makecell[tc]{Average $l_{1}$ norm of \\invalid filters}&\makecell[tc]{Accuracy on \\test set} \\ 
			\hline
			\multirow{4}{*}{CIFAR-10}&
			\multirow{2}{*}{0.1}
			&baseline&6.982&0.003&93.64\\
			&&grafting+&\textbf{12.117}&\textbf{1.385}&\textbf{94.30}\\
			\cline{2-6}
			&\multirow{2}{*}{1}
			&baseline&7.035&0.095&93.64\\
			&&grafting+&\textbf{12.191}&\textbf{1.635}&\textbf{94.30}\\
			\hline
			\multirow{4}{*}{CIFAR-100}&
			\multirow{2}{*}{0.1}
			&baseline&12.086&0.001&71.26\\
			&&grafting+&\textbf{18.064}&\textbf{0.654}&\textbf{73.37}\\
			\cline{2-6}
			&\multirow{2}{*}{1}
			&baseline&12.126&0.075&71.26\\
			&&grafting+&\textbf{18.119}&\textbf{0.842}&\textbf{73.37}\\
			\hline
		\end{tabular}\\
	\end{center}
	\label{table:branches_2}
\end{table*}

\subsection{Evaluating the Grafting+}\label{sec_4_4}

\textbf{grafting+ on CIFAR-100}: We compare grafting+ with the baseline, knowledge distillation and filter grafting on CIFAR-100 datasets. The training setting of the baseline, knowledge distillation and filter grafting is introduced in Section \ref{sec_4_3}. For grafting+, we use one teacher network and two student networks in this experiment. Knowledge distillation is performed at each optimization step while grafting is performed at the end of each epoch. At the end of the training, we select one student network for testing. For each method on each dataset, we do five runs and report the $mean$ and $std$ of the accuracy. The results are shown in Table \ref{table:grafting+_cifar}. We can see that grafting+ gives the best results on CIFAR-100 dataset. 

It is worth noting that the setting in this experiment is different with Table \ref{table:4_3}. In Table \ref{table:4_3}, the networks in grafting has different learning rate to increase diversity. However, in Table \ref{table:grafting+_cifar}, to prove grafting+ can outperform vanilla distillation, the teacher always has better performance than the student, and the learning rate schedule of grafting is consistent with vanilla distillation in this experiment. But we can still observe that the performance of grafting+ on CIFAR-100 is better than grafting in Table \ref{table:4_3}. 

\begin{figure}[!h]
	\centering
	\includegraphics[width=15cm,]{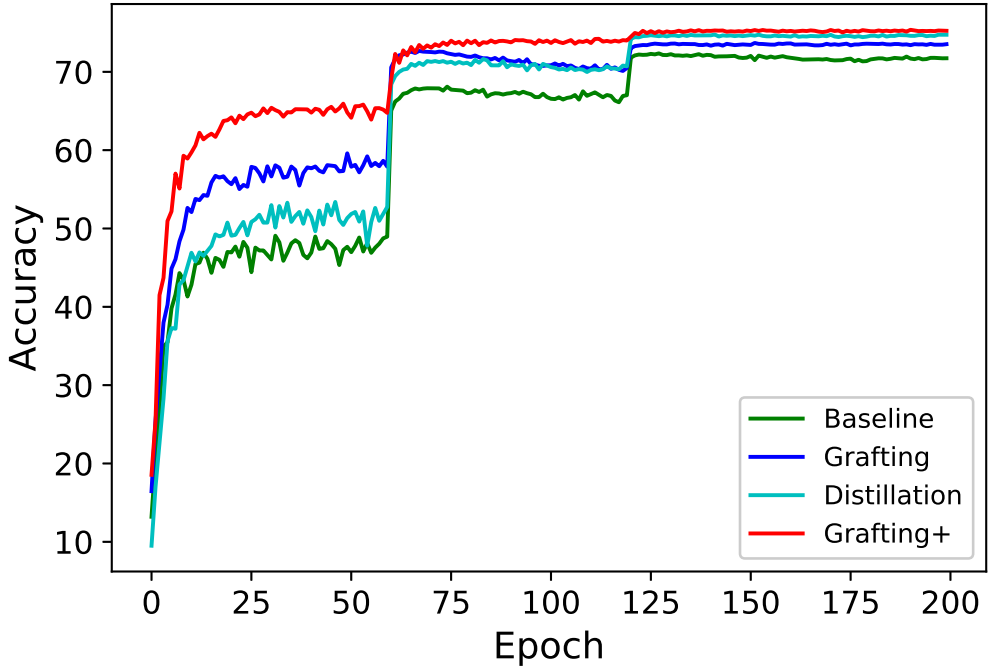}
	\caption{This figure depicts the validation accuracy of each methods on CIFAR-100 datasets. The network is MobileNetV2. (best viewed in color)}
	\label{figure: accuracy_curve}
\end{figure}

We further depict the accuracy with the training epochs of each method in Figure \ref{figure: accuracy_curve}. It's interesting that grafting mostly influences the model at the early training epochs while distillation mostly makes an impact at later stages. Also, the network trained by grafting+ achieves the best accuracy among all the methods. 

\textbf{Extending grafting+ to multiple networks}: In previous experiment, grafting+ only considers one teacher network and two student networks. However, this is the simplest form of this framework. We find that grafting+ could further improve the network by bringing more teachers and students to the framework. In this experiment, we use ResNet-110 as the teacher and MobileNetV2 as the student. The model accuracy with the number of teachers and students in grafting+ is listed in Table \ref{table:number_T_S}.  It can be found that as we raise the number of teachers and students, the model accuracy increases. The results have shown that given more networks, there exists great potential for the grafting+ framework.

\begin{table}[!h]
	\caption{This table records the student accuracy with number of teachers and students in grafting+ framework. The network structures for teacher and student are ResNet-110 and MobileNetV2, respectively.}
	\begin{center}
		\begin{tabular}{|c|c|c|} 
			\hline
			Dataset&Setting&Test accuracy\\
			\hline
			\multirow{5}{*}{CIFAR-100}& teacher*1, student*2& 75.18 $\pm$ 0.12 \\
			& teacher*1, student*4& 75.45 $\pm$ 0.02 \\
			& teacher*1, student*6& 76.85 $\pm$ 0.02 \\
			& teacher*2, student*2 & 75.96 $\pm$ 0.33\\
			& teacher*3, student*2 & 76.44 $\pm$ 0.29\\
			\hline
		\end{tabular}
	\end{center}
	\label{table:number_T_S}
\end{table}

\begin{figure}[!t]
	\centering
	\includegraphics[width=15cm,]{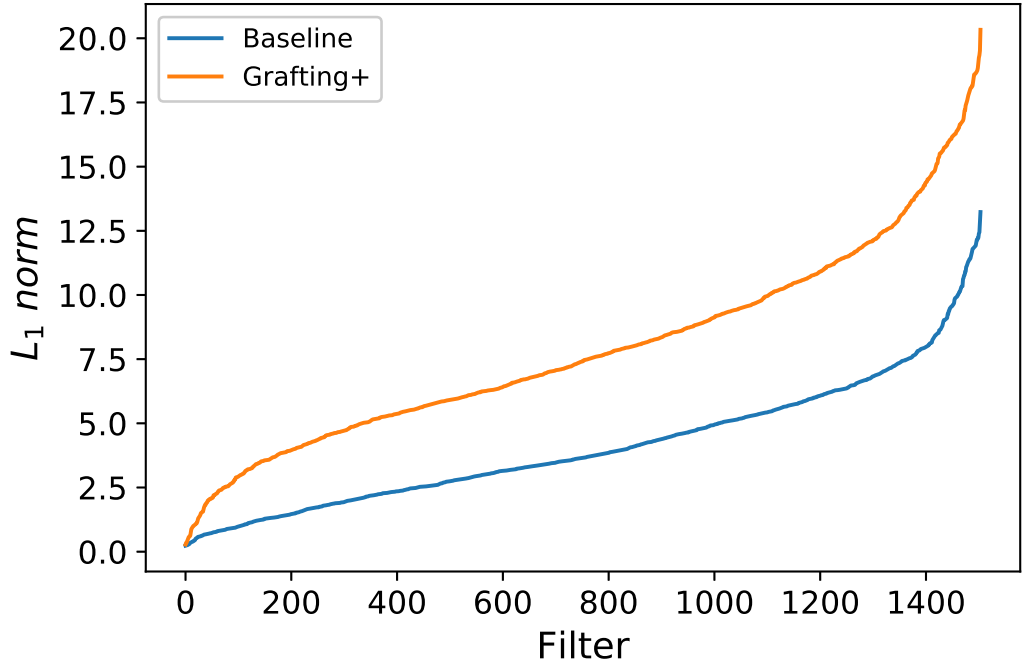}
	\caption{This figure shows the filters' $l_{1}$ norm of Conv3 layer for baseline and grafting+. (best viewed in color)}
	\label{figure:filter_norm}
\end{figure}

\textbf{Evaluating grafting+ in the filter level}: 
In order to show that the network trained by grafting+ does improve both valid and invalid filters, we calculate the average $l_{1}$ norm of the valid and invalid filters in Table \ref{table:branches_2} and plot all the filters' $l_{1}$ norm in Figure \ref{figure:filter_norm}. The results show that grafting+ could greatly densify both valid and invalid filters given a fixed model structure. As we state that distillation and grafting are complementary in the filter level, their combination could boost a filter-efficient network. 

There are more experiments in the experiments including ablation study on hyper-parameters, more comparison results and visual understanding of grafting+ framework.

\section{Conclusion}
In this work, a new learning paradigm called ‘filter grafting’ is proposed. We argue that there are two key points for effectively applying filter grafting algorithm: 1) how to choose proper criterion to calculate the inherent information of filters in DNNs. 2) how to balance the co-efficients of information among networks. To deal with these two problems, we propose entropy-based criterion and adaptive weighting strategy to increase the network’s performance. In addition, we find that grafting and distillation has complementary effect on filters. Thus we propose grafting+ framework to improve both valid and invalid filters. Heuristically, there are some future directions to be considered: 1) how to apply grafting on multiple networks with different network structures; 2) how to improve the network’s performance with larger number of networks in grafting+ framework.

\small
\bibliographystyle{unsrt}
\bibliography{references}

\end{document}